\documentclass{article}
\usepackage{style}

\usepackage{soul}
\usepackage{times}
\usepackage[T1]{fontenc}
\usepackage{hhline}
\usepackage{algorithm}
\usepackage{natbib}
\usepackage[noend]{algorithmic}
\usepackage{soul}
\usepackage{url}
\usepackage[hidelinks]{hyperref}
\usepackage[utf8]{inputenc}
\usepackage[small]{caption}
\usepackage{graphicx}
\usepackage{amsmath}
\usepackage{multirow}
\usepackage{amsthm}
\usepackage{booktabs}
\usepackage{xspace}
\usepackage{amssymb}
\usepackage{url}
\usepackage{pgf, tikz}
\usetikzlibrary{arrows, automata}
\usepackage{tabularx}
\usepackage{subfig}
\usepackage{mdframed}

\newtheorem{definition}{Definition}
\newtheorem{example}{Example}
\newtheorem{proposition}{Proposition}

\newcommand{\prompt}[1]{\mathsf{P}^{#1}}

\def\<{\langle}
\def\>{\rangle}
\def\-{\mbox{-}}
\usepackage{mathrsfs}

\newcommand{\textt}[1]{\mbox{$\tt #1$}}

\newcommand{\q}{q}
\newcommand{\ABox}{\ensuremath{\mathcal{A}}}
\newcommand{\TBox}{\ensuremath{\mathcal{T}}}

\def\DLL{\ensuremath{\mi{DL\-Lite}}\xspace}

\def\DLLR{\ensuremath{\mi{DL\-Lite_{\mathcal{R}\xspace}}}\xspace}

\def\={\,\text{=}\,}

\usepackage[most]{tcolorbox}

\tcbset{
  boxstyle/.style={
    colback = cyan!2,
    colframe= -red!85!green!50,
    arc=1.5mm,
    boxrule=1pt,
    left=1mm, right=1mm,
    top=1mm, bottom=1mm,
    fonttitle=\bfseries\sffamily,
    coltitle=white,
    title={#1},
  }
}

\newtcolorbox{BoxPrompt}[2][]{%
  boxstyle={#2},%
  #1%
}

\newtcolorbox{BoxText}[2][]{%
  boxstyle={#2},%
  #1%
}

\definecolor{tagbg}{RGB}{255,229,154}
\definecolor{tagtxt}{RGB}{255,140,0}

\definecolor{bluebg}{RGB}{154,228,255}
\definecolor{bluetxt}{RGB}{20,110,230}

\definecolor{tealbg}{RGB}{154,243,238}
\definecolor{tealtxt}{RGB}{0,160,150}

\newcommand{\tagbox}[1]{%
  \begingroup
  \setlength{\fboxsep}{1pt}%
  \colorbox{tagbg}{\textcolor{black}{\textbf{#1}}}%
  \endgroup
}

\newcommand{\bluehl}[1]{%
  \begingroup
  \setlength{\fboxsep}{1pt}%
  \colorbox{bluebg}{%
    \hspace{1pt}\textcolor{black}{\textbf{#1}}\hspace{1pt}%
  }%
  \endgroup
}
\newcommand{\tealhl}[1]{%
  \begingroup
  \setlength{\fboxsep}{1pt}%
  \colorbox{tealbg}{%
    \hspace{1pt}\textcolor{black}{\textbf{#1}}\hspace{1pt}%
  }%
  \endgroup
}

\newcommand{\boon}[1]{{\tt #1}}

\newcommand\ta[1]{\ensuremath{{\tt TA}_{#1}}\xspace}
\def\t{{\color{black}w}}

\def\wrt{w.r.t.\xspace}

\newcommand{\Set}[1]{\ensuremath{#1}\xspace}

\def\arg{\Set{A}}
\def\att{\Set{R}}
\def\supp{\Set{S}}

\def\QBAF{\ensuremath{\Lambda}\xspace}

\def\FAKB{\ensuremath{{\cal K}}\xspace}

\def\nQBAF{QBAF\xspace}
 \def\nFAKB{FAKB\xspace}

\newcommand{\tQBAF}{\ensuremath{\< \arg,\att,\supp,\tau\>}\xspace}
 \newcommand{\tFAKB}{\ensuremath{\< \TBox\!,\ABox\>}\xspace}

\def\DLL{\mbox{\it DL-Lite}}
\def\DLLR{\mbox{\it DL-Lite$_{\cal R}$}}

 \def\LM{\textsf{M}}

\usepackage{xcolor}
\usepackage{adjustbox}
 \usepackage[switch]{lineno} 
\usepackage{booktabs}
\usepackage{xcolor}
\usepackage{tikz}
\usepackage{array}

\definecolor{cSup}{RGB}{210, 245, 210}
\definecolor{cAtt}{RGB}{255, 215, 215}
\definecolor{cNei}{RGB}{240, 240, 240}
\definecolor{cEmp}{RGB}{255, 255, 255} 

\newcommand{\splitcell}[5]{%
    \begin{tikzpicture}[baseline=(current bounding box.center)]
        \def\w{1.6cm}
        \def\h{0.9cm}
        
        \fill[#1] (0,\h) -- (\w,\h) -- (0,0) -- cycle;
        \fill[#3] (0,0) -- (\w,0) -- (\w,\h) -- cycle;
        
        \draw[white, thick] (0,0) -- (\w,\h);
        
        \node[anchor=north west, inner sep=2pt] at (0,\h) {\tiny \textbf{#2}};
        
        \node[anchor=south east, inner sep=2pt] at (\w,0) {\tiny \textbf{#4}};
        
        \draw[black, thick, #5] (0,0) rectangle (\w,\h);
    \end{tikzpicture}%
}

\def\tit{LLM-based Argument Mining meets Argumentation and Description Logics: a Unified Framework for Reasoning about Debates}

\author{
Gianvincenzo Alfano\and
Sergio Greco\and
Lucio La Cava\and\\
Stefano Francesco Monea\and
Irina Trubitsyna\\
\affiliations
Department of Informatics, Modeling, Electronics and System Engineering\\
University of Calabria, Italy\\
\emails
\{g.alfano, greco, lucio.lacava, trubitsyna\}@dimes.unical.it;\ stefanofrancesco.monea@unical.it
}

\begin{document}
\title{\tit}
\maketitle

\begin{abstract} 
Large Language Models (LLMs) achieve strong performance in analyzing and generating text, yet they struggle with explicit, transparent, and verifiable reasoning over complex texts such as those containing debates. 
In particular, they lack structured representations that capture how arguments support or attack each other and how their relative strengths determine overall acceptability. 
We encompass these limitations by proposing a framework that integrates learning-based argument mining with quantitative reasoning and ontology-based querying. 
Starting from a raw debate text, the framework extracts a fuzzy argumentative knowledge base, where arguments are explicitly represented as entities, linked by attack and support relations, and annotated with initial fuzzy strengths reflecting plausibility \wrt the debate's context.
Quantitative argumentation semantics are then applied to compute final argument strengths by propagating the effects of supports and attacks. 
These results are then embedded into a fuzzy description logic setting, enabling expressive query answering through efficient rewriting techniques. 
The proposed approach provides a transparent, explainable, and formally grounded method for analyzing debates, overcoming purely statistical LLM-based analyses.
\end{abstract}

\section{Introduction}\label{sec:intro}

Large Language Models (LLMs) have recently demonstrated impressive capabilities in understanding, translating, and generating natural language across a wide range of domains.
They are able to summarize, identify stances, and even generate persuasive arguments. 
However, despite these successes, LLMs have well-known limitations when it comes to reasoning tasks, which are often implicit, opaque, and difficult to verify.
These limitations are particularly critical in analyzing political debates, where transparency, explainability, and coherence of reasoning are essential.

A key reason for these shortcomings is that LLMs operate as statistical learners over large text corpora. 
While they can implicitly encode argumentative discourse patterns, they lack an explicit representation of argumentative structure, such as which claims attack or support others, how strong these relationships are, and how the balance of a debate should be assessed. 
As a consequence, LLM-based analyses are hard to inspect, cannot be systematically queried, and do not naturally support formal consistency checking.

An LLM can summarize a debate or generate a balanced narrative. 
However, if asked ``what are the most acceptable left-wing arguments on climate?'', the answer is typically qualitative, non-repeatable, and unsupported by an explicit line of reasoning, thus lacking explainability. 
To overcome these limitations, there is a growing consensus that learning-based approaches should be integrated with symbolic and sub-symbolic reasoning frameworks. 
In particular, Argument Mining techniques have emerged as a promising class of methods for extracting structured representations of arguments from text, providing a crucial bridge between unstructured natural language and structured representations amenable to formal reasoning.
The aim is to identify arguments and relationships such as attacks and supports, often with associated confidence or strength values.
A toy example about climate change policies is presented next.

\begin{figure*}[t]
    \centering  \includegraphics[width=\linewidth]{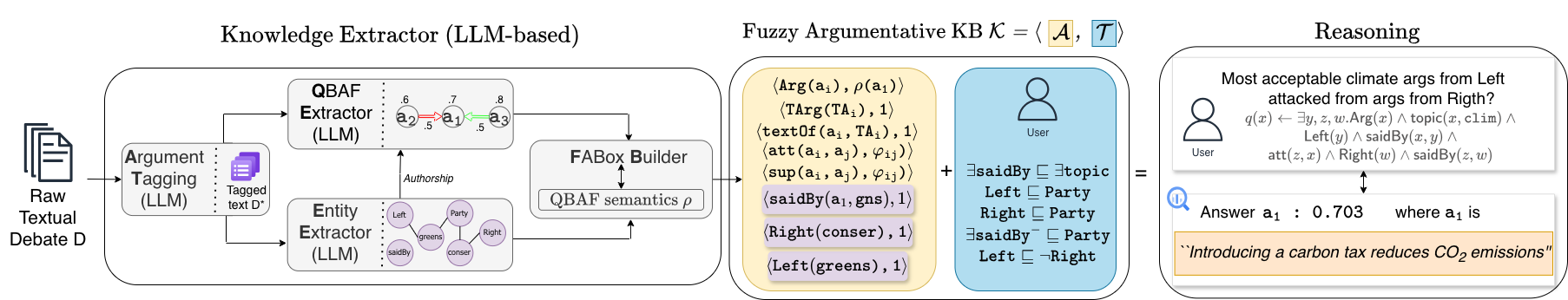}
    \caption{\nFAKB workflow: from raw unstructured texts to \nFAKB generation and querying.}
    \label{fig:architecture}
\end{figure*}

\begin{example}\label{ex:intro0}\rm
Consider the debate ($D$) on climate change.

\begin{mdframed}[backgroundcolor=gray!10, topline=false, bottomline=false, rightline=false, linewidth=3pt, linecolor=gray]\footnotesize
Representatives of the \underline{greens}, traditionally associated with the \underline{left}, argued that (${\tt a}_1$) ``\textit{Introducing a carbon tax reduces $CO_2$ emissions}''. 
They emphasized that the \underline{party} endorses this argument as a central element of its \underline{climate} agenda.
In response, \underline{conservatives} expressed strong opposition, claiming that (${\tt a}_2$) ``\textit{A carbon tax increases costs for households and firms}'', reflecting economic concerns and motivating their rejection of the proposed climate measures.
Further, the  \underline{greens} stated that (${\tt a}_3$) ``\textit{Revenues from a carbon tax can be invested in renewable energy}''.
\end{mdframed}
Using argument mining, the debate can be seen as a set of arguments $
\arg = \{{\tt a}_1, {\tt a}_2, {\tt a}_3\}$, and the set of the following argumentative relations: $\{ {\tt a}_2$ attacks ${\tt a}_1$, ${\tt a}_3$ supports ${\tt a}_1 \}$.

Arguments can be associated with an initial strength through a function $\tau$, reflecting factors such as the plausibility \wrt the context of the debate, e.g.,
$\tau_{{\tt a}_1} = 0.7$, $\tau_{{\tt a}_2} = 0.6$, and $\tau_{{\tt a}_3} = 0.8$. 
Similarly, an initial strength is also defined for attacks and supports. 
The final framework is shown in Figure~\ref{fig:architecture}, where attacks (resp., supports) are represented by red-colored (resp., green-colored) arrows.\hspace*{\fill}~$\Box$
\end{example}

While the argument-based representation adopted in the previous example allows for a more transparent view of the debate, it still does not answer the question of how the overall balance of attacks and supports affects the final acceptability of each argument.
Indeed, the final strength of argument ${\tt a}_1$ should, on the one hand, increase as it is supported by argument ${\tt a}_3$, and, on the other hand, decrease, as it is attacked by argument ${\tt a}_2$. 
Thus, it is necessary to update the strength of arguments taking into account how arguments attack and support each other. 
This can be carried out by means of a suitable semantics for Quantitative Bipolar Argumentation Framework (QBAF)~\citep{cayrol2005graduality,baroni2019fine,amgoud2018gradual,Mossakowski-core}.

Generally, a concrete argument is associated with a sentence with a proper meaning and with several additional information that could be very important in answering queries and making informed decisions.
By modeling a problem through a plain \nQBAF, we may neglect important information that can be very useful, such as those concerning the author, the text of the argument, the date it was introduced, the topic (e.g., what the argument is about), the polarity (e.g., whether the sentence conveys a positive, negative or neutral sentiment), the polarity rating, the sentence style (e.g., veracity, sarcasm, irony), the city and nationality of the author, and much more.
Therefore, there is a need for expressive yet tractable mechanisms to integrate them with background knowledge and to answer complex queries with graded outcomes. 
That is, the argumentation framework should be coupled with a Knowledge Base (KB) containing further information contained in the text and related to arguments.

In our running example, several relevant pieces of information are neglected when considering only the level of abstraction provided by a \nQBAF, such as the topic (e.g., Climate) or the name and wing of political parties (which we have underlined in the text for clarity).

We propose a novel framework called \textit{Fuzzy Argumentative Knowledge Base} (\nFAKB), that combines learning-based extraction with quantitative reasoning and ontology-based query answering. The framework is organized as a pipeline of distinct modules (see  Figure~\ref{fig:architecture}). 
First, an extractor—potentially based on state-of-the-art argument mining techniques—processes a debate and produces an argumentative knowledge base that subsumes a \nQBAF. 
The extracted knowledge is represented as a fuzzy \DLL\ ABox, called (source) \textit{fuzzy argumentative} ABox (FABox), where arguments are explicitly modeled as individuals, linked by attack and support relations, and annotated with initial strengths reflecting the \textit{plausibility} w.r.t. the debate's context. Unlike traditional methods that rely on external heuristics to assign initial strengths, our framework also consider transformer's output log-probabilities, capturing the model's internal confidence during the extraction phase.
The extracted structure (i.e., the source FABox) is then interpreted as a \nQBAF to compute final argument strengths via established gradual semantics. This step is crucial because initial strengths obtained during extraction mainly reflect plausibility; the \nQBAF semantics make the debate's implicit balance explicit by propagating the impact of attackers and supporters in a principled, explainable manner.

After extracting the FABox (where arguments, attacks, and supports have associated fuzzy values, whereas the remaining entities and relationships are crisp), the user adds their own knowledge through a TBox consisting of a set of crisp \DLL\xspace axioms, useful for checking the consistency of the extracted data and {for building a \nFAKB, combining debate's information with user's general knowledge, to be queried later}.
The answer to a (relational calculus) query consists of a set of fuzzy tuples, where each tuple is associated with a confidence degree.

\vspace*{2mm}
\noindent
\textbf{Contributions.} 
We propose the \emph{Fuzzy Argumentative Knowledge Base} (\nFAKB), a novel framework that integrates learning-based argument extraction with symbolic reasoning, quantitative argumentation semantics, and ontology-based query answering. Specific contributions are:

\begin{list}{$\bullet$}
     {\setlength{\topsep}{0mm}
		  \setlength{\rightmargin}{+0mm}
	    \setlength{\leftmargin}{+2.5mm}
      \setlength{\itemindent}{0mm}
			\setlength{\itemsep}{-0.0mm}}
\item A pipeline that extracts a \emph{fuzzy argumentative ABox} (FABox) from raw debate text using LLM-based argument mining, where arguments, attacks, and supports are represented as \DLL\xspace fuzzy concept and role instances.

\item 
A new method for assigning strengths to arguments, attacks, and supports.
Rather than directly adopting the initial strengths from prompting strategies—a common practice in current literature—our framework improves these values by taking into account the transformer's output {log-probabilities}. 
This captures the model's internal confidence during the extraction phase. Argument strength is then refined via established gradual \nQBAF semantics.

\item 
We qualitatively and quantitatively compare our method with existing prompt-based solutions, showing a more me-aningful distribution of argument and relation strengths.

\item 
 We equip the \nFAKB with a reasoning mechanism to support consistency checking, and formal query answering returning \emph{fuzzy tuples} with associated confidence degrees.
\end{list}
\vspace*{1mm}

Moving beyond simple argumentative queries on the final acceptability of a single goal argument, our framework supports complex (conjunctive) queries over both built-in argumentative predicates (e.g., \textt{att}, \textt{sup}) and the broader extracted knowledge (e.g., discourse topics, polarity, etc).

To the best of our knowledge, this is the first work formally
integrating LLM-based argument mining, quantitative argumentation semantics, 
and fuzzy ontology-mediated reasoning, overcoming core limitations of purely 
learning-based models and providing an explainable and 
queryable methodology for reasoning over complex (textual) debates.

\section{Preliminaries}\label{sec:prel}

After recalling the Quantitative Bipolar Argumentation Framework and basic concepts underlying Description Logics, we briefly review the main Argument Mining tasks.

\subsection{Quantitative Bipolar AF}\label{subsec:QBAF}

A Quantitative Bipolar Argumentation Framework (\nQBAF) $\QBAF=\tQBAF$ consists of a set of arguments $\arg$, {disjoint} sets of attack relations $\att\subseteq \arg\times\arg$ and support relations $\supp\subseteq \arg\times\arg$, and a total function $\tau: \arg\rightarrow [0,1]$ that assigns the initial strength $\tau_{a}$ to every argument $a\in\arg$. 
Thus, a \nQBAF can be represented by a graph $\<\arg,\att,\supp\>$ with two types of edges 
denoting the relation $a_i$ attacks/supports $a_j$. 
Intuitively, attacking arguments contribute to decreasing the strength of the attacked argument, while supporting arguments contribute to increasing the strength of the supported argument.
Since the strength of arguments is a real value in the range $[0,1]$, a strength $0$ corresponds to the total rejection of an argument, $1$ to the total acceptance, and values between $0$ and $1$ express various degrees of acceptance.

\smallbreak

\noindent
\textbf{Semantics.}
Several gradual semantics have been proposed so far to compute the acceptance of arguments. 
That is, a gradual semantics is the way a total \emph{update function} $\rho$, assigning a value in $[0,1]$ to all arguments, is computed. 
\nQBAF semantics are typically classified within the family of \textit{modular semantics}~\citep{Mossakowski-core}, consisting in applying in sequence $(i)$ an aggregation function $\alpha$, taking into account the strength of attackers and supporters, and $(ii)$ an influence function, taking into account the value computed by $\alpha$ and the initial strength of the argument.
The final strength $\rho_{\QBAF}(a)$ (or simply $\rho(a)$) of argument $a\in\arg$ is iteratively computed, starting from  $\tau_{a}$, by considering the strength $\rho({a_j})$ of each attacker/supporter $a_j$ of $a$.

We next recall one (among several existing) modular semantics, called \textit{Quadratic Energy}~\citep{Potyka18QE}. 
This semantics uses the following \textit{sum-based} aggregation function:

\begin{equation}\label{eq:sum}
\resizebox{.71\linewidth}{!}{
$\displaystyle
\alpha(a_i)=\sum\limits_{(a_j,a_i)\in\supp}\rho(a_j)-\sum\limits_{(a_j,a_i)\in\att}\rho(a_j)
$}
\end{equation}

\noindent 
while, let $E(a_i) = \frac{\alpha(a_i)^2}{1 + \alpha(a_i)^2}$, the update function is:
\begin{equation}\label{eq:qen}
\resizebox{.80\linewidth}{!}{$\rho(a_i)=
\begin{cases}
\left(1 - E(a_i)\right)\cdot \tau_{a_i}  & \text{if } {\alpha(a_i)} \le 0 \\
E(a_i) + \left(1 - E(a_i)\right)\cdot \tau_{a_i}  & \text{if } {\alpha(a_i)} > 0 
\end{cases}$}
\end{equation}

For acyclic \nQBAF, $\rho$ is computed in linear time following a topological ordering of the arguments~\citep{Potyka19}.

\begin{example}\label{ex:prel-qbaf}\rm
Consider the \nQBAF $\QBAF =\tt \< \{{\tt a}_1, {\tt a}_2, {\tt a}_3\},$ $\tt \{({\tt a}_2,{\tt a}_1)\},$ $\{({\tt a}_3,{\tt a}_1)\}, \{\tau_{{\tt a}_1}={0.7},\tau_{{\tt a}_2}=0.6,\tau_{{\tt a}_3}=0.8\} \>$ underlying Example~\ref{ex:intro0}.
As $\alpha({\tt a}_1) {=} \tt 0.2$, then $E({\tt a}_1) \approx 0.039$ and $\rho({\tt a}_1) = E({{\tt a}_1}) + \left(1 - E({{\tt a}_1})\right)\cdot \tau_{{\tt a}_1}\approx\tt 0.712$. 
\hspace*{\fill}$\Box$
\end{example}

\subsection{Description Logics}\label{subsec:DL}

A DL \emph{knowledge base} (KB), hereafter simply called a knowledge base, consists of an {\emph{Assertional Box}} ABox (database) and a {\emph{Terminological Box}} TBox (ontology), which are constructed from a set $\mathsf{N_C}$ of \emph{concept names} (unary predicates), a set of $\mathsf{N_R}$ of \emph{role names} (binary predicates), and a set $\mathsf{N_I}$ of \emph{individuals} (constants).
Specifically, an \emph{ABox} is a finite set of \emph{concept  assertions} of the form $\textt{A}(a)$, where $\textt{A} \in \mathsf{N_C}$ and $a \in \mathsf{N_I}$, and \emph{role assertions} of the form $\textt{r}(a, b)$, where $\textt{r} \in \mathsf{N_R}$ and $a, b \in \mathsf{N_I}$. 
A TBox is a finite set of axioms whose form depends on the considered DL. 
In $\DLLR$, TBox axioms are only \emph{concept inclusions} of the form $\textt{B} \sqsubseteq \boon{C}$ and role inclusions of the form $\tt Q \subseteq S$, formed using the following syntax (where $\textt{A} \in \mathsf{N_C}$, $\textt{r} \in \mathsf{N_R}$, and $\textt{r}^-$ is the inverse of role $\textt{r}$):
{\setlength{\abovedisplayskip}{2pt}
\setlength{\belowdisplayskip}{2pt}
\begin{align*}
\textt{B} := \textt{A} \mid \exists \textt{Q}; && \textt{Q} := \textt{r} \mid \textt{r}^-;&&
\textt{C} := \textt{B} \mid \neg \boon{B}; &&  \boon{S} := \textt{Q} \mid \neg \textt{Q}.
\end{align*}}
An inclusion is said \emph{positive} (resp. \emph{negative}) whenever the right-hand side contains a positive (resp. \emph{negated}) element.

Several variants of \DLL\ have been proposed.
Hereafter we refer to \DLLR, which provides the logical underpinnings of the OWL 2 QL~\citep{OWLProfiles}, though our results apply to all query-rewritable \DLL\ fragments.~\citep{CalvaneseGLLR07,Lenzerini-AAAI20,Bienvenu16,ArtaleCKZ09,BienvenuO15}.

\smallbreak 

\noindent
\textbf{Semantics.} 
An \emph{interpretation} is of the form $\mathcal{I} = (\Delta^\mathcal{I}, \cdot^\mathcal{I})$, 
where $\Delta^\mathcal{I}$ is a non-empty set called \emph{interpretation domain} and $\cdot^\mathcal{I}$ is an \emph{interpretation function} mapping each $\textt{A} \in \mathsf{N_C}$ to $\textt{A}^\mathcal{I} \subseteq \Delta^\mathcal{I}$, each $\textt{r} \in \mathsf{N_R}$ to $\textt{r}^\mathcal{I} \subseteq \Delta^\mathcal{I} \times \Delta^\mathcal{I}$, and each $a \in \mathsf{N_I}$ to $a^\mathcal{I} \in \Delta^\mathcal{I}$, with $a \neq b$ implying $a^\mathcal{I} \neq b^\mathcal{I}$ (the last condition is the well-known Unique Names Assumption).
Being $\cdot^\mathcal{I}$ a function, it can be defined as follows:
$i)$ $\boon{A}^{\cal I}:\Delta^{\cal I} \rightarrow \{0,1\}$ with 
$(\boon{A}(a))^{\cal I} = 1$ if $a^{\cal I} \in \boon{A}^{\cal I}$, $0$ otherwise; and 
$ii)$ $\boon{Q}^{\cal I}:\Delta^{\cal I} \times \Delta^{\cal I} \rightarrow \{0,1\}$, with $(\boon{Q}(a,b))^{\cal I} = 1$ if $(a^{\cal I},b^{\cal I}) \in \boon{Q}^{\cal I}$, $0$ otherwise. 
The function is extended to general concepts and roles as follows: 
$(i)$ $(\exists \textt{Q}(a))^\mathcal{I} = 1$ if $\exists b . (\boon{Q}(a,b))^{\cal I} = 1$, $0$ otherwise; 
$(ii)$ $(\neg \boon{A}(a))^\mathcal{I} = 1 - (\boon{A}(a))^\mathcal{I}$;
$(iii)$ $(\textt{Q}^-(a,b))^\mathcal{I} = (\boon{Q}(b,a))^{\cal I}$, 
$(iv)$ $(\neg \textt{Q}(a,b))^\mathcal{I} = 1 - (\textt{Q}(a,b))^\mathcal{I}$.
An interpretation $\mathcal{I}$ \emph{satisfies} an inclusion $\textt{G} \sqsubseteq \textt{H}$ if $(\textt{H}(a))^\mathcal{I}\geq (\textt{G}(a))^\mathcal{I}$ for any $a\in {\mathsf{N_I}}$;
it satisfies $\textt{A}(a)$ (resp. $\textt{r}(a, b)$) if $(\textt{A}(a))^\mathcal{I} \geq 1$ (resp., $(\textt{r}(a,b))^\mathcal{I} \geq 1$). 
$\mathcal{I}$ is a \emph{model} of $\tFAKB$ iff $\mathcal{I}$ satisfies all inclusions in $\TBox$ and all assertions in $\ABox$. 
If a KB has a model, it is said to be \textit{consistent}, otherwise, it is said to be inconsistent.

\begin{example}\label{ex:prel-DLLITE}\rm
Consider the following knowledge base.
\smallskip 

\hspace*{-7mm}
\begin{tabular}{cc}
\begin{minipage}[t]{0.7\columnwidth}
\small
$\ABox_{\ref{ex:prel-DLLITE}}\!:\! \begin{cases} 
\textt{saidBy}(\textt{a}_1, \textt{g}),\\ \textt{saidBy}(\textt{a}_2, \textt{c}),\\ \textt{saidBy}(\textt{a}_3, \textt{g}),\\ \textt{Left}(\mathtt{g}), \textt{Right}(\textt{c}),\\
\textt{topic}(\textt{a}_i,\textt{climate})\mid  i\!\in\![1..3]
\end{cases}$
\end{minipage}
&
\hspace*{-18mm}
\begin{minipage}[t]{0.6\columnwidth}
\small
$\TBox_{\ref{ex:prel-DLLITE}}\!:\!\! \begin{cases} 
(1)\ \exists\textt{saidBy} \sqsubseteq\exists \textt{topic}\\
(2)\ \textt{Left} \sqsubseteq \textt{Party}\\
~~~~~\ \textt{Right} \sqsubseteq \textt{Party} \\
~~~~~\ \exists\textt{saidBy}^{-} \sqsubseteq \textt{Party} \\
(3)\ \textt{Left} \sqsubseteq \neg\textt{Right}
\end{cases}$
\end{minipage}
\end{tabular}

\smallskip
\noindent 
The interpretation $\mathcal{I}$ with $\Delta^\mathcal{I}=\{\textt{a_1, a_2, a_3, g, c}\}$, $\textt{saidBy}^\mathcal{I} = \{(\textt{a_1, g}), (\textt{a_2, c}), (\textt{a_3, g})\}$, $\textt{Left}^\mathcal{I}=\{\textt{g}\}$, $\textt{Right}^\mathcal{I}=\{\textt{c}\}$, $\textt{Party}^\mathcal{I}=\{\textt{c},\textt{g}\}$, and $\textt{topic}^\mathcal{I}=\{(\textt{a_i}, \textt{climate})\mid i\in[1..3]\}$ is a model of $\<\ABox_{\ref{ex:prel-DLLITE}},\TBox_{\ref{ex:prel-DLLITE}}\>$.
\hfill $\Box$
\end{example}

A conjunctive query (CQ)  is a formula $\exists \vec{y}\,\varphi(\vec{x},\vec{y})$ where $\varphi$ is a conjunction of atoms.  
A boolean CQ has no free variables $\vec{x}$ and is entailed by a KB if it holds in all its models.  
A tuple $\vec{t}$ of individuals is a \emph{certain answer} if substituting the free variables with $\vec{t}$ yields a boolean CQ entailed by the KB.  
A union of CQs (UCQ) is a disjunction of CQs; its answers are the union of the answers of its disjuncts.

\subsection{Argument Mining and LLMs}
\label{subsec:argmining}
Argument Mining (AM) is the research area concerned with the automatic identification and extraction of argumentative structures from text~\citep{lawrence-reed-2019-argument}. Given a document or a dialogue, the goal is to detect argumentative units—such as claims, premises, and evidence—and to identify their relationships, most notably attacks and supports~\citep{Cabrio2018FiveYO}. 
AM provides a crucial interface between raw text and structured reasoning frameworks, and it has been deeply studied in contexts such as political debates, legal documents~\citep{Palau2009argumentation}, scientific articles~\citep{sukpanichnant2024peerarg}, and online discussions~\citep{habernal-gurevych-2017-argumentation}.

We assume the existence of a finite set $\mathcal{V}$ of words called \textit{vocabulary}, and a concatenation operator $\circ$ taking sequences of words. 
For the sake of presentation, we consider words as the fundamental units of analysis, rather than conventional tokens.   
A textual document is a sequence $D$$=$$[w_1,\dots,w_n]$ of words in $\cal V$.
A \textit{textual argument} of $D$ is any subsequence $\ta{i}$$=$$[\t_{b_i}, ..., \t_{e_i}]$ of $D$, consisting of one or more sentences to be classified according to predefined labels. 
The AM problem consists of four (sub)problems:

\begin{list}{-}
     {\setlength{\topsep}{0mm}
		  \setlength{\rightmargin}{+0mm}
	    \setlength{\leftmargin}{+2.5mm}
      \setlength{\itemindent}{0mm}
			\setlength{\itemsep}{-0.5mm}}
\item 
\textit{Argument Segmentation}: chunks a document $D$ into a sequence of non-overlapping, contiguous \textit{partitional arguments} $\textt{PA}_1,\dots\textt{PA}_k$, such that $D=\textt{PA}_1\circ \cdots \circ \textt{PA}_k$; 
\item
\textit{Argument Classification}: assigns a class label (e.g., premise, claim, major claim) to each textual argument identified in $D$ after the segmentation;

\item
\textit{Relation Identification}: determines whether a directed argumentative relation holds between any pair of (labeled) textual arguments in $D$, following the previous steps;

\item
\textit{Relation Classification}:  establishes the type (if any) of argumentative relation between pairs of textual arguments, according to predefined types (e.g., support, attack).
\end{list}

\smallbreak

\noindent
\textbf{Large Language Models.}\  
We denote by $\LM_\theta$ any pretrained Large Language Model (LLM) where $\theta$ may correspond to the set of parameters of a general-purpose or task-specific (i.e., fine-tuned) LLM. 
$\LM_\theta$ (or simply $\LM$ whenever $\theta$ is understood) corresponds to a decoder-only Transformer architecture, which maps any input (possibly raw) sequence of words $D=[w_1,\dots,w_k]$ into an output sequence of words $D'=[w_{k+1},\dots,w_n]$ by iteratively generating each $w_i = \LM([w_1,\dots,w_{i-1}])$ (with $i \in [k\mbox{+}1,n]$).
Basically, $\LM$ is a classifier that assigns a probability distribution over the vocabulary $\mathcal{V}$, reflecting the likelihood of each word being generated, after a sequence $D$ has been given as input. 
To this end, we denote by $\LM(D) = [\Pr(w_1|D),$ $\dots,$ $\Pr(w_{m}|D)]$ the probability distribution over the $m$ words of the vocabulary $\cal V$, that will be used to select the next word.
Given the typically large size of the vocabulary, probabilities $\Pr(\cdot)$ are often very low. 
To prevent numerical underflow problems, the natural logarithm is applied to them, obtaining the so-called \textit{log-probabilities} $\ell(w_i|D) = \mathrm{ln}\big(\Pr(w_i|D)\big)$.


\section{Reasoning on Textual Dialogues}\label{sec:architecture}

Recent research increasingly emphasizes the integration of LLMs with formal reasoning frameworks~\citep{FreedmanAAAI25}.
LLMs represent a promising approach to transform natural language debates into structured representations. 

We adopt this perspective and provide an LLM-based architecture (shown in Figure~\ref{fig:architecture}) that extracts, from a raw text representing a debate, a knowledge graph containing the relevant information such as arguments, relations, and additional concepts considered relevant w.r.t. the debate.
The resulting knowledge will then be {stored} in a fuzzy {set of facts} called \textit{source fuzzy argumentative ABox} (FABox), where uncertainty is modeled in terms of fuzzy concepts and roles. 
The FABox, updated with the final argument strength computed via a gradual semantics over the underlying \nQBAF, is then paired with a TBox encoding general domain knowledge expressed by the user. The result is a \textit{Fuzzy Argumentative Knowledge Base} (\nFAKB),  presented next.

While classical logic relies on absolute true/false opposites, fuzzy logic maps truth onto a continuous spectrum~\citep{Zadeh65}. 
It has been integrated into several fields, e.g. databases~\citep{chen2012fuzzy,petry2012fuzzy,yazici1992survey,yazici2013fuzzy}, answer-set programming~\citep{NieuwenborghCV07,BlondeelSVC13}, description logics~\citep{Lukasiewicz08,PasiP23}, and formal argumentation~\citep{wu2016godel}.
 
Our framework extends fuzzy \DLL\ by also encoding argument-based concepts and roles. 
To this end, we assume the existence of: 
$(i)$ an additional enumerable, linearly ordered set $\mathsf{N_A} = \{a_1,a_2,...\}$ of distinguished individuals used to refer to arguments (identifiers), and $(ii)$ the built-in concept names 
$\textt{Arg}$ and $\textt{TArg}$, respectively taking values from $\mathsf{N_A}$ and $\mathsf{N_T}$.
Finally, we assume existence of a set of roles $\mathsf{N_F}$ (argument features) taking values from 
$\mathsf{N_A} \times \mathsf{N_I}$, and of the built-in role names $\textt{att}$, $\textt{sup}$, and $\textt{textOf}$, where the first two take values in $\mathsf{N_A} \times \mathsf{N_A}$, whereas the last one takes values in $\mathsf{N_A} \times \mathsf{N_T}$.
We use the symbols $\mathsf{N_I^+} \!=\! \mathsf{N_I}\! \cup\! \mathsf{N_A}\!\cup\! \mathsf{N_T}$, $\mathsf{N_C^+} \!=\! \mathsf{N_C}\! \cup\! \{\textt{Arg}\}\!\cup \!\{\textt{TArg}\}$ and $\mathsf{N_R^+}\! =\! \mathsf{N_R}\! \cup\! \mathsf{N_F}\! \cup\! \{\textt{att},\textt{sup}\}\cup \{\textt{textOf}\}$
to denote the \textit{augmented} sets of individuals, concepts, and roles, respectively.

We also extend, as done in \citep{Potyka21MLP}, the \nQBAF introduced in Section~\ref{sec:prel} to model strengths also on attacks and supports. 
Hereafter a \nQBAF is a tuple $\<\arg,\att,\supp,\tau,\varphi\>$ where $\varphi: \att \cup \supp \rightarrow [0,1]$ is a total function assigning a strength (i.e., $[0,1]$-value) to attacks and supports.
With respect to the framework introduced in Section~\ref{sec:prel}, the semantics is defined by only changing the definition of the aggregation function, e.g. by replacing $\rho(a_j)$ in Equation~\ref{eq:sum} with $\rho(a_j) \times \varphi(a_j,a_i)$.

\begin{example}\label{ex:QBAF-with-attacks}\rm
Let $\QBAF'$ be the \nQBAF obtained from the \nQBAF $\QBAF$ of Example~\ref{ex:prel-qbaf} by adding $\varphi(\textt{a_2},\textt{a_1})$ ${=}$ $\varphi(\textt{a_3},\textt{a_1}){=}1/2$. As $\alpha(\textt{a_1})\mbox{=}0.1$,  the final strength of $\textt{a_1}$ under Quadratic Energy (cf. Equation~\ref{eq:qen}) is $\rho({\textt{a_1}})\approx 0.703$.\hspace*{\fill}$\Box$
\end{example}

We next present the Fuzzy Argumentative ABox, storing the fuzzy knowledge extracted from (raw, textual) debates.

\begin{definition}\label{def:fabox}\rm
A \emph{Fuzzy Argumentative ABox} (FABox) is a finite set 
of fuzzy concepts $\<\textt{A}(a),v\>$ and role assertions $\<\textt{r}(a, b),v\>$, with 
$(i)$ $\textt{A} \in \mathsf{N_C^+}$, 
$(ii)$ $\textt{r} \in \mathsf{N_R^+}$, and 
$(iii)$ $v\!\in\![0,1]$ iff $\textt{A}\=\textt{Arg}$ or $\textt{r}\!\in\! \{\textt{att}, \textt{sup}\}$, $v\=1$\  \text{otherwise}.
\end{definition}

Condition $(iii)$ states that fuzzy degrees can only be associated with arguments, attacks, and supports. 
Clearly, for concepts and roles, the domain restrictions previously introduced continue to hold.
As it will be clarified in the next sections, this fuzziness intuitively captures the \textit{plausibility} in the extraction process, and will be used to compute the final strength of arguments. 
That is, this special treatment reserved for arguments only is due to the fact that the attack and support interactions among arguments were not taken into account during the extraction phase.

\begin{definition}\label{def:fabox-consistent}\rm
A FABox $\ABox$ is said to be \textit{consistent} iff:
\begin{itemize}
\item 
$\<\textt{Arg}(a),u\>$, $\<\textt{Arg}(a),v\>$ $\in \ABox$ implies $u=v$; 
\item 
$\<\textt{r}(a_i,a_j),u\>$, $\<\textt{s}(a_i,a_j),v\>$ $\in \ABox$ with $\textt{r,s}\in \{\textt{att,sup}\}$ implies $ \textt{r}\mbox{=}\textt{s}$,  $u\mbox{=}v$, and $\<\textt{Arg}(a_i),v_i\>$, $\<\textt{Arg}(a_j),v_j\>\in \ABox$;
\item \textt{textOf} defines a bijection between arguments identifiers and textual arguments in $\ABox$; and $\<\textt{textOf}(a_i,\textt{TA}_{i}),1\>\in \ABox$  iff $\<\textt{Arg}(a_i),v_i\>$, $\<\textt{TArg}(\textt{TA}_{i}),1\>\in \ABox$. 
\end{itemize}
\end{definition}
Intuitively, the first condition states that an argument can have at most one strength associated with it. Then, argument concepts must belong to the FABox whenever they appear in any attack or support role, and there cannot exist two distinct relations from any pair of arguments.
Finally, the last condition ensures that any argument must have exactly one text associated with it. 
The motivation for introducing consistent FABoxes is, as stated next, that they subsume a \nQBAF.
Before presenting the result, we introduce some notation.
For any concept $\textt{A}$ and role $\textt{r}$ of $\ABox$, we denote with 
$\textt{A}^{\!1}$=$\{ a \mid \< \textt{A}(a), v\> \in \ABox \}$, 
$\textt{A}^{\!2}$=$\{ (a,v) \mid \< \textt{A}(a), v\> \in \ABox \}$, 
$\textt{r}^{1}$=$\{ (a,b) \mid \< \textt{r}(a,b), v\> \in \ABox \}$, and 
$\textt{r}^{2}$=$\{ ((a,b),v) \mid$ $\< \textt{r}(a,b), v\> \in \ABox \}$ the projection on individuals and degrees. 

\begin{proposition}\label{prop:subsumes-qbaf}
For any consistent FABox $\ABox$, {$\QBAF_{\!_\ABox}=\<\textt{Arg}^{1},\textt{att}^{1},\textt{sup}^{1},  \textt{Arg}^{2},  \textt{att}^{2}\cup\textt{sup}^{2}\>$} forms a \nQBAF.
\end{proposition}

Hereafter, we assume that the FABox $\ABox$ is consistent, and simply call $\QBAF_\ABox$ the \textit{underlying} \nQBAF w.r.t. $\ABox$.

\begin{example}\label{ex:fabox0}\rm
A possible consistent source FABox, extracted from debate $D$ of Example~\ref{ex:intro0}, is $\ABox^{\rm s}\mbox{=}\{\<x,1\>\mid x\in \ABox_{\ref{ex:prel-DLLITE}}\} \cup$ $\ABox_{\ref{ex:fabox0}}$ where $\ABox_{\ref{ex:prel-DLLITE}}$ is the ABox discussed in Example~\ref{ex:prel-DLLITE}, and

\noindent
$\ABox_{\ref{ex:fabox0}} = \begin{cases} 
    \<\textt{Arg}(\textt{a_1}), 0.7\>, \<\textt{Arg}(\textt{a_2}), 0.6\>, \<\textt{Arg}(\textt{a_3}), 0.8\> \\
    \<\textt{att}(\textt{a_2, a_1}), 0.5\>, \<\textt{sup}(\textt{a_3, a_1}), 0.5\>\\
    \<\textt{TArg}(\ta{a_i}), 1\>\ \text{with}\ i\in[1..3]\\
    \<\textt{textOf}(\textt{a_i}, \ta{a_i}), 1\>\ \text{with}\ i\in[1..3]
\end{cases}$
\noindent
with $\ta{a_i}$\!\! be the italicized textual arguments
in Example~\ref{ex:intro0}.~$\Box$
\end{example}

The component of our architecture that is in charge of extracting the source FABox $\ABox^{\rm s}$ is called \textit{Knowledge Extractor} (see Figure~\ref{fig:architecture}). 
As it will be further explained in the next Section, it employs LLM-based argument mining techniques, some of which contain technical novelty. Specifically, the extraction of initial strengths, rather than relying on purely prompting strategies, also takes into account the log-probability of the transformer's output, capturing the model's internal confidence during extraction.  
These strengths are refined by applying a gradual semantics over the underlying \nQBAF $\QBAF_{\ABox^\text{\rm s}}$ to account for inter-argument relationships not captured during the initial extraction phase.
This process yields an \textit{updated} FABox, which is then paired with a user-defined TBox to form the following framework.

\begin{definition}\rm
A \emph{Fuzzy Argumentative KB} (\nFAKB) is a pair $\FAKB=\tFAKB$ where $\ABox$ is an (updated) consistent FABox, and $\TBox$ is a {\DLLR}\ TBox such that built-in concept and role names  (i.e. \textt{Arg, TArg, att, sup} and \textt{textOf}) do not occur in the right-hand side of inclusions.
\end{definition}

The restriction on the form of assertions states that arguments, attacks, and supports can only be defined extensively by facts contained in $\ABox$, that is, they are only extracted from the input debate. 
Clearly, a {\DLLR\xspace} KB is a special case of \nFAKB, where elements in the ABox hold with degree $v=1$.

\begin{example}\label{ex:FAKB-syntax}\rm
A possible \nFAKB, extracted from $D$, is $\FAKB = \<\ABox_{\ref{ex:FAKB-syntax}},\TBox_{\ref{ex:prel-DLLITE}}\>$ where $\ABox_{\ref{ex:FAKB-syntax}}$ is obtained from the (source) FABox $\ABox^{\rm s}$ of Example~\ref{ex:fabox0} by updating initial argument strengths, and the user-defined knowledge $\TBox_{\ref{ex:prel-DLLITE}}$ is that of Example~\ref{ex:prel-DLLITE},  where: axiom~(1)  intuitively states that arguments must have a topic. Axioms in~(2) specify that any entity endorsing an argument is a political party, in turn constituted by both Left and Right parties. 
Finally, axiom~(3) enforces the mutual exclusivity between left-wing and right-wing parties.\hspace*{\fill}$\Box$
\end{example}

\section{Extracting FABox through LLMs}
The wide variety and quantity of data on which LLMs are pre-trained give them a solid knowledge base, enabling them to tackle knowledge extraction tasks even under limited data availability.  
As shown in Figure \ref{fig:architecture}, we leverage such capabilities to build a Knowledge Extractor module as the core of our proposed framework. The activities performed by the Knowledge Extractor consist of the following four tasks:  
$i)$ Argument Tagging, $ii)$ Entity Extraction, $iii)$ QBAF Extraction, and $iv)$ FABox Builder.
We next discuss the four tasks. 

\subsection{Argument Tagging}

This task consists of identifying in the input text the arguments of the debate and separating them from the rest of the text, which defines the informative context. However, the standard \textit{Argument Segmentation} problem (AS) described in Section~\ref{subsec:argmining} is not sufficient for this purpose, as it does not account for non-argumentative textual spans. 

For this reason, we resort to the double-tag approach of~\citep{caputo2026eacl}, which explicitly separates argumentative from non-argumentative content, enabling the subsequent phases of our framework. The approach makes use of a task-specific LLM $\LM$, which has been fine-tuned for solving AS, and extensively validated against human annotations. $\LM$ acts as a mapping function that applied to $\prompt{tag}\<D\>$, where let $\prompt{tag}$ being a specific prompt template, $\prompt{tag}\<D\>$ denotes the prompt filled with $D$, gives as output a new document $D^*$ derived from $D$ by adding pairs of tags ($\<\textt{AC}_i\>$, $\</\textt{AC}_i\>$), with $i\in\{1,2,..,k\}$ delimiting (textual) arguments. An example is presented next.
 
\begin{example}\label{ex:tag}\rm    
Consider the debate $D$ of Example~\ref{ex:intro0}. The tagged version $D^*=\LM(\prompt{tag}\<D\>)$ is as follows.

\begin{mdframed}[backgroundcolor=gray!10, topline=false, bottomline=false, rightline=false, linewidth=3pt,innerleftmargin=5pt, linecolor=gray]\footnotesize
Representatives of the greens $\cdots$ argued that 
\bluehl{\it $\<\textt{AC}_1\>$Introducing a carbon tax reduces $CO_2$ emissions$\</\textt{AC}_1\>$} $\cdots$ 
In response, conservatives expressed $\cdots$ that   \tealhl{\it $\<\textt{AC}_2\>$ A} \tealhl{\it carbon tax increases costs for households and firms$\</\textt{AC}_2\>$},~$\cdots$
Further, the greens stated that 
\tagbox{\it $\<\textt{AC}_3\>$ Revenues  from a carbon} \tagbox{\it  tax can be invested in renewable energy $\</\textt{AC}_3\>$}.\hspace*{\fill}$_\Box$
\end{mdframed} 
\end{example}

Hence, document $D^*$ is given as input to the Entity and QBAF Extractor tasks, which are presented next.

\subsection{Entity Extractor}\label{sec:entity-extractor}

This two-phase task extracts, 
in the form of a knowledge graph, explicit information from the tagged document, such as the entities appearing in the debate, the subjects they refer to, and the relationships that hold among them. 

The first phase, referred to as \textit{entity extraction}, aims at identifying specific entities (e.g., distinct people and political parties such as \textit{John} and \textit{Greens}) present in the tagged document, i.e., the portion of text that is not part of the textual arguments identified in the tagging phase. 
Through a prompt $\prompt{ner}$, we use an LLM $\LM$ as a mapping function that applied to $\prompt{ner}\< D^*\> $: $(i)$ determines a set of entities extracted from $D^*$, $(ii)$ classifies the previously obtained set of entities into general classes or concepts (e.g., \textit{John} and \textit{Greens} are respectively classified as  \textit{Politician} and \textit{LeftParty}), and $(ii)$ then generates concept instances (e.g., \textit{Politician(John)} and \textit{LeftParty(Greens)}).

The second phase, called \textit{role extraction}, detects the contextual and interpersonal links connecting such entities, extracted analogously to the entity and concept names. After having identified a set of binary relation names among entities (e.g. \textit{memberOf}), we follow a few-shot approach, as found promising in~\citep{PolatTG25}, to ask the LLM to identify, through a prompt {$\prompt{kb}$}, a set of semantic relations connecting entities to entities (e.g. \textit{memberOf(John, Greens)}) or textual arguments to entities. In the latter case (e.g. \textit{John} authored $\ta{i}$), a role instance connecting the corresponding argument identifier to the entity is generated (e.g. \textit{author($a_i$,John)}).

\subsection{QBAF Extractor}

This task takes as input a tagged text and gives as output a QBAF-based KB, where each argument (identifier) is associated with the related textual argument. 

We use $\ta{i}^*=\<\textt{AC}_i\> \ta{i} \</\textt{AC}_i\>$ to denote the (i-th) \textit{textual argument} in $D^*$. 
The task starts by associating a unique \textit{argument (identifier)} $a_i \in \mathsf{N_A}$ to textual arguments $\ta{i} \in D^*$, following the sequential ordering of arguments in $D^*$ (recall that $a_1 \!\!<\!\! a_2 \!\!<\!\!a_3\!\!<\!\! \cdots$). 
Thus, for each textual argument $\ta{i}$, it introduces the concept instances $\textt{TArg}(\ta{i})$ and $\textt{Arg}(a_i)$ and the role instance $\textt{textOf}(a_i,\ta{i})$. 
For instance, on the tagged document $D^*$ in Example~\ref{ex:tag}, it yields $\textt{Arg}(a_i)$ and $\textt{textOf}(a_i, \ta{i})$ with $i\in [1..3]$, where e.g. $\ta{1}\=$\textit{``Introducing a carbon tax reduces $CO_2$ emissions''}.

\vspace*{1mm}
\noindent
\textbf{Argument Strength.}
Unlike existing approaches where the initial argument strength is obtained by simply prompting LLMs~\citep{FreedmanAAAI25}, we also consider the \textit{argument plausibility}, expressed as the probability that the textual argument will be generated by the transformer, based on the text of the debate preceding the current argument.

Specifically, the intrinsic plausibility of each argument $\ta{i}=[\t_{b_i},\dots,\t_{e_i}]$ is obtained by looking at `how much' the LLM itself considers coherent the generation of $\ta{i}$  \wrt the text in the document that precedes  $\ta{i}$.
As words in the arguments are generated sequentially, for each word $\t_j$ in $\ta{i}$, the text in $D$ that precedes $w_j$ (i.e., $[\t_1,\dots,\t_{j-1}]$) has to be considered. Formally: 
{\setlength{\abovedisplayskip}{1pt}
\setlength{\belowdisplayskip}{1pt}
\begin{equation}\label{eq:logits-unscaled}
    \mu_{i} = \frac{1}{1+(e_i{-}b_i)}\sum_{j=b_i}^{e_i} \ell(\t_j \mid [\t_1,\dots,\t_{j-1}]).
\end{equation}
}

Intuitively,  the higher $\mu_{i}$, the more the text $\ta{i}$ (and thus the argument $a_i$) is considered \textit{plausible}---simply put, $\ta{i}$ contains words that the model considers expected \wrt the preceding text in $D$.

The reason to operate on log-probabilities $\ell(\cdot)$ is to preserve the original model preferences, making $\mu$ values suitable for relative comparison across arguments. 
As they range in the interval $(-\infty,0]$, they are normalized in the $[0,1]$-interval as follows:
{\begin{equation}\label{eq:logits-score-norm}
\overline{\mu}_{i}=\frac{\mu_{i}-\mathrm{min}(\boldsymbol{\mu})}{\mathrm{max}(\boldsymbol{\mu})-\mathrm{min}(\boldsymbol{\mu})},
\end{equation}}

\noindent
where  $\boldsymbol{\mu}=\{\mu_{i}\mid i\in [1,k]\}$ and $\overline{\boldsymbol{\mu}}=\{\overline{\mu}_{i}\mid i\in [1,k]\}$.
Then, let $\tau_{a_i}^p$ (or simply $\tau_{i}^p$) be the initial strength of $a_i$ obtained by prompting an LLM $\LM$, the following initial strength is assigned to $a_i$:
{\begin{equation}\label{eq:logits-score}
    \tau_{i}=\mathrm{max}\big(0,\mathrm{min}(1,\tau_{i}^p\cdot \big(1+\overline{\mu}_{i}-\overline{\boldsymbol{\mu}}_{\rm avg}\big)\big))
\end{equation}

\noindent
where $\overline{\boldsymbol{\mu}}_{\rm avg}$ is the average of elements in $\bar{\boldsymbol{\mu}}$.
Intuitively, when $\overline{\mu}_{i}=\overline{\boldsymbol{\mu}}_{\rm avg}$, the plausibility of $a_i$ is aligned with the mean across all arguments, and thus we neither penalize nor reward the prompted value, i.e. $\tau_{i}=\tau_{i}^p$. 
Differently, $\overline{\mu}_{i}>\overline{\boldsymbol{\mu}}_{\rm avg}$ (resp. $\overline{\mu}_{i}<\overline{\boldsymbol{\mu}}_{\rm avg}$) states that $a_i$ is more (resp. less) plausible than the average and, thus, we increment or decrement $\tau_{i}^p$ with a value equal to 
$\tau_{i}^p\cdot (\overline{\mu}_{i}-\overline{\boldsymbol{\mu}}_{\rm avg})$. 
As the updated values can exceed the $[0,1]$ interval, a clamping function $\mathrm{max}(0,\mathrm{min}(1,\cdot))$ is applied.

As discussed in Section~\ref{sec:validation}, our approach achieves a more meaningful strength distribution than existing methods.

We next present the procedure to extract argumentative relations from the previously identified arguments.

\vspace*{1mm}
\noindent
\textbf{Attack/Support Extraction.}\ 
Since we are also interested in associating strengths to relations, we extend the promising LLM-based strategy proposed in~\citep{gorur-etal-2025-large}. That is, similarly to the initial argument strength extraction, we propose to obtain argumentative relations and their strength by looking at the LLM's internals.

Before explaining our approach, we introduce some notation. 
Let $\ta{i}^*$ be a tagged textual argument from $D^*$, we use $\mathit{Aut}_{a_i}$ to denote the author of argument $a_i$, received from the Entity Extraction. 
Moreover, we use $D^*_{\leq i}$ to denote the portion of $D^*$ from the beginning to argument $\ta{i}^*$ (included).  

For each pair $(\ta{i}, \ta{j})$ of textual arguments with $i > j$, we leverage an LLM $\LM$ as a classification tool that applied to $\prompt{rel}\<D^*_{\leq i}, \mathit{Aut}_{a_i},\ta{i}, \mathit{Aut}_{a_j},\ta{j}\>$ gives as output a probability distribution considering the relation types in the restricted vocabulary ${\cal V}'=\{\mathit{Attack}$, $\mathit{Support}$, $\mathit{Unrelated}\}$.
Note that, relation strengths are determined dynamically based only on the information available at the moment the argument $a_i$ is introduced (i.e., based on $D^*_{\leq i}$).

Note that, as we are referring to argumentative debates, the motivation for choosing $i > j$ is that arguments may attack or support previous arguments only.
To ensure a proper probability distribution over relation types, we first extract the log-probabilities 
$\ell(r~\mid~\prompt{rel}\<D^*_{\leq~i},~\mathit{Aut}_{a_i},\ta{i}, \mathit{Aut}_{a_j},\ta{j}\>)$
and then apply the softmax function to them, obtaining $p_{i,j}^{r}$.

Intuitively, $p_{i,j}^{r}$ represents the model's confidence, considering the tagged document $D^*$, that the relation from $a_{i}$ to $a_{j}$ is of type $r$.
The final decision on the corresponding relation is hence identified by the following process. Given a threshold $\vartheta \in [0,1]$, and let $\{x,y,z\} = \cal V'$ with $x \in \{\mathit{Attack}, \mathit{Support}\}$, we assign to the relation from $a_i$ to $a_j$ the type $x$ whenever the following inequality holds
{\setlength{\abovedisplayskip}{3pt}
\setlength{\belowdisplayskip}{3pt}
\begin{equation}\label{eq:prob_label_relation_diseq}
    p_{i,j}^{x} - (p_{i,j}^{y} +  p_{i,j}^{z}) > \vartheta
\end{equation}}

\noindent
and, in such a case, we associate the strength value $\varphi_{i,j}=p_{i,j}^{x} {-} (p_{i,j}^{y} {+}  p_{i,j}^{z})$ to it. Intuitively, a relation type $x$ from $a_i$ to $a_j$ exists if the model's confidence in $x$ exceeds the combined confidence in the other two types by at least $\vartheta$.

We observe that this procedure should not be regarded as an ultimate solution for identifying argumentative relations.
Rather, it represents a simple and effective heuristic based on the next-token probabilities produced by the LLM.

\subsection{FABox Builder}\label{subsec:extr3}
The aim of this task is to integrate, in a structured way, the output of the QBAF Extractor (i.e., the QBAF-based knowledge graph) with that of the Entity Extractor (i.e., the knowledge graph representing concepts and roles).
To this end, it starts by populating a source FABox $\ABox^{\rm s}$ with fuzzy assertions of the form $\<\textt{Arg}(a_i),\tau_{a_i}\>$, $\<\textt{TArg}(\ta{i}),1\>$, $\<\textt{textOf}(a_i,\ta{i}),1\>$, $\<\textt{att}(a_i,a_j),\varphi_{i,j}\>$, and $\<\textt{sup}(a_i,a_j),\varphi_{i,j}\>$, received from the QBAF-Extractor.

Then, any other concept $\textt{C}(a)$ and role $\textt{r}(a,b)$ extracted by the Entity Extractor is stored in $\ABox^{\rm s}$ under the form of fuzzy element $\<\textt{C}(a),1\>$ and $\<\textt{r}(a,b),1\>$.

As in assigning the initial argument strength the attacks and supports weight has not yet been taken into account; updated argument strengths are computed by calling some gradual semantics on the QBAF $\QBAF_{\ABox^{\rm s}}$ underlying the FABox $\ABox^{\rm s}$. That is, any fuzzy argument assertion $\<\textt{Arg}(a),\tau_{a}\>$ is replaced with $\<\textt{Arg}(a),\rho({a})\>$, where $\rho$ is the update function under the chosen gradual semantics. 
The result is an (updated) FABox $\ABox$, an example of which, derived from a more complex version of our running example, is shown in the form of a knowledge graph in Figure~\ref{fig:step3}.

It is worth noting that, although the exploration of gradual semantics lies beyond the scope of this work, our results are orthogonal to the choice of gradual semantics.

\begin{proposition}\label{prop:LLM}
Let $D$ be a document, and $\LM_\theta$ an LLM. Then, the updated FABox corresponding to $D$ and $\LM_\theta$ can be computed in polynomial time \wrt the number of words in $D$.
\end{proposition}

\begin{figure}
    \centering
    \includegraphics[width=\linewidth]{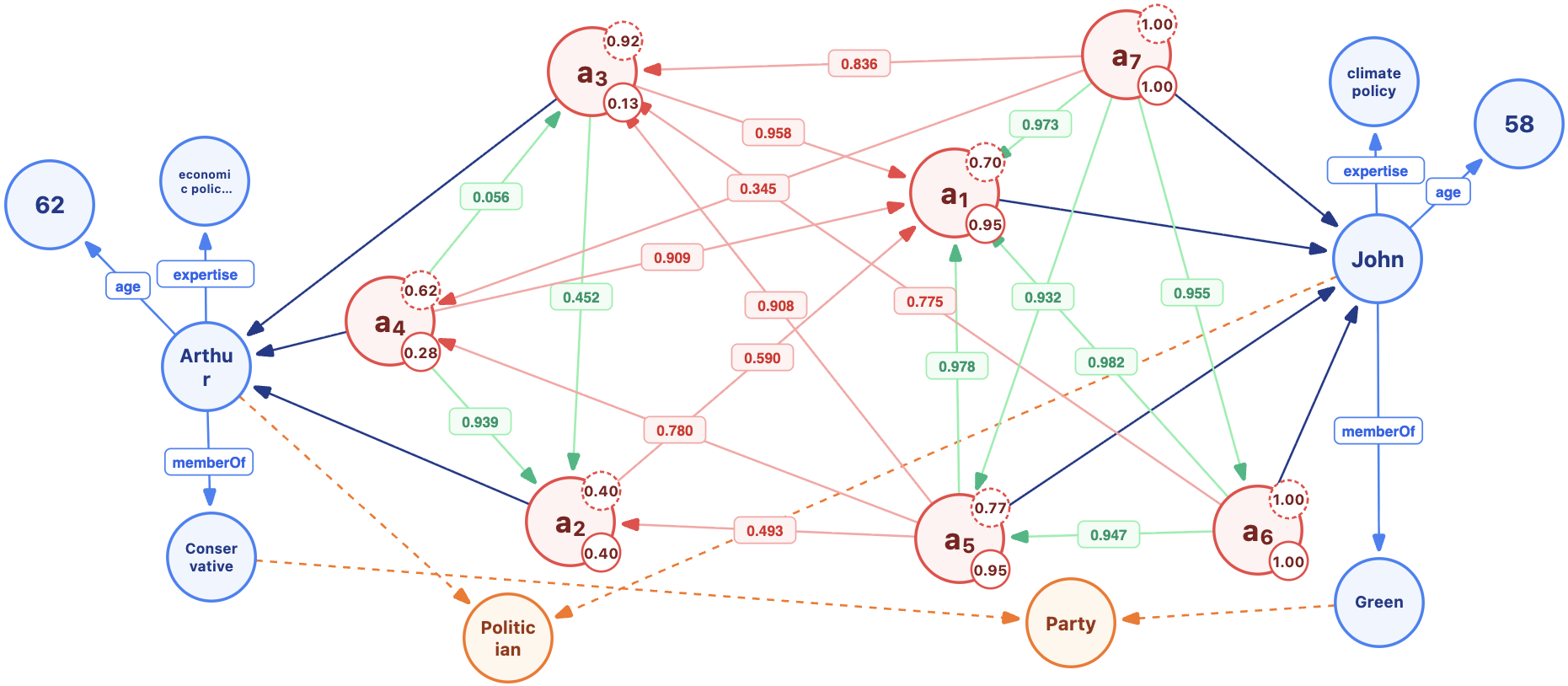}
    \caption{An example of updated FAbox, visualized in the form of a knowledge graph, and extracted from an extended version of debate $D$ of Example~\ref{ex:intro0}. Nodes are color-coded by type: blue for entities, red for argument identifiers, and yellow for concepts. Similarly, edges represent specific relations: green/red denotes support/attacks, while dark blue, light blue, and yellow represent authorship, role instances, and concept instances, respectively. Upper/lower numbers within each argument denote its initial/updated strength. For optimal visualization, the arguments' text has been omitted, yet they can be found in Section~\ref{sec:validation}. A full-screen visualization also including additional entities and full texts is provided in the Appendix.}
    \label{fig:step3}
\end{figure}

\section{Validation}\label{sec:validation}

We next validate one of the novelties of our proposal, namely the computation of the initial strengths for arguments, attacks, and supports, obtained by also considering the log-probabilities (cf. Equations~(\ref{eq:logits-unscaled})-(\ref{eq:prob_label_relation_diseq})), and compare them with pure prompt-based counterparts.
Since the debate $D$ of Example~\ref{ex:intro0} is too simplistic, we created a set ${\cal D}=\{D_1,\dots D_{10}\}$ of ten distinct debates on different topics, each consisting of at least 289 words (323.5 on average). 
During Argument Tagging, we obtained an average of 9.9 distinct arguments using the SoTA LLM-based model of (Caputo et al. 2026)~\nocite{caputo2026eacl}---which achieves $94\%$ correlation with human annotators and an $F_1$ score of $0.92$.
For all remaining tasks, we employed the pretrained LLM \textt{Qwen2.5\ 7B}. 
A more extensive parametric analysis regarding the choice of LLM is deferred to future work.
Moreover, the Entity Extractor led to the creation of 14.8 entities, 6.1 concept instances, and 13 (non-argumentative) relations among entities, averaged on all debates in $\cal D$. 

\noindent\textbf{Argument Strength.}
We compare the proposed argument strength estimation (cf. Equation~\ref{eq:logits-score}) and its prompt-based counterpart. 
The latter, as in~\citep{FreedmanAAAI25}, assigns the initial strengths $\tau_i$ by prompting an LLM without looking at its internals and specifically asking a value in $[0,1]$.

The following table reports the identifier, the text, and the initial strength of the arguments extracted from $D_1$, which is an extended version of $D$ consisting of {$327$} words (instead of {$80$} in $D$). For clarity, argument identifiers are superscripted by author: either $(A)$rthur or $(J)$ohn.

{\centering
\scriptsize
\renewcommand{\arraystretch}{0.95}
\setlength{\tabcolsep}{0pt}
\begin{tabular}{>{\centering\arraybackslash}p{0.7cm}p{5.7cm}>{\centering\arraybackslash}p{1.1cm}|>{\centering\arraybackslash}p{1.1cm}}
\toprule
\centering 
\multirow{1}{*}{\textbf{$a_i^{A\!/\!J}$\ \ }} & \multirow{1}{*}{\textbf{Textual Argument ($\ta{i}$) in $D_1$}} & 
\textbf{Our} & \textbf{Prompted}\\
\midrule

$a_1^J$ & \textit{`introducing a carbon tax reduces CO2 emissions'} & $.70$ & $.9$\\
\cline{2-4}
\multirow{2}{*}{$a_2^A$} & \textit{`we cannot ignore the immediate economic reality for} & \multirow{2}{*}{$.40$} & \multirow{2}{*}{$.9$}\\
& \textit{the people of Birmingham and beyond'} & &\\

\cline{2-4}

$a_3^A$ & \textit{`a carbon tax increases costs for households and firms'} & $.92$ & $.8$ \\

\cline{2-4}
\multirow{2}{*}{$a_4^A$} & \textit{`at a time when inflation is high, adding this financial} &\multirow{2}{*}{$.62$} & \multirow{2}{*}{$.8$}\\ 
&\textit{burden motivates our rejection of the proposed measures} & &\\ 

\cline{2-4}
  
$a_5^J$ & \textit{`this isn't money lost;it's money repurposed for our future'} & $.77$ & $.8$ \\
\cline{2-4}
\multirow{2}{*}{$a_6^J$} & \textit{`We have clearly stated that revenues from a carbon tax} 
& \multirow{2}{*}{$1.$} & \multirow{2}{*}{$.8$} \\
&\textit{can be invested in renewable energy'} &&\\
\cline{2-4}

\multirow{2}{*}{$a_7^J$} & \textit{`This creates a cycle where the tax funds the very solution} & \multirow{2}{*}{$1.$} & \multirow{2}{*}{$.8$} \\
&\textit{that will lower energy costs in the long run'} & &\\

\midrule

& \textbf{Avg. $\pm$ Std. Dev.} on all args in $D_1$ &  $\mathbf{.77\! \pm\! .2}$ & $\mathbf{.83\! \pm\! .05}$ \\
\bottomrule

\end{tabular}
}

In the prompt-based approach, plausibility is treated as an independent variable for each argument---a trend evidenced by the consistently high scores and low variance. Indeed, it can be observed that the prompt-based approach assigns a more flattened score to almost all textual arguments {($\approx 0.83 \pm 0.05$)}, thus preventing any meaningful differentiation among arguments and demonstrating less sensitivity to plausibility. 
In contrast, our approach yields a differentiated and more semantically interpretable distribution of plausibility values {($\approx 0.77 \pm 0.20$)}, with arguments that are more (resp., less) central or contextually coherent receiving higher (resp., lower) scores.
This qualitative analysis is also confirmed across all datasets in $\cal D$. 
Indeed, on average, our approach yields a distribution of ${0.74 \pm 0.25}$, contrasting the flatter behavior of the prompt-based one (i.e., {$0.76 \pm 0.11$}), thus indicating a limited ability to differentiate between strong and weak arguments.

\smallbreak

\begin{table}[t!]
    \centering
    \renewcommand{\arraystretch}{0.5}
    \setlength{\tabcolsep}{3pt}
    
    \renewcommand{\splitcell}[5]{%
    \begin{tikzpicture}[baseline=(current bounding box.center)]
        \def\halfw{0.7cm} 
        \def\h{0.7cm}     
        
        \fill[#1] (0,0) rectangle (\halfw,\h);
        \node[font=\footnotesize\bfseries] at (0.5*\halfw, 0.5*\h) {#2};
        
        \fill[#3] (\halfw,0) rectangle (2*\halfw,\h);
        \node[font=\footnotesize\bfseries] at (1.5*\halfw, 0.5*\h) {#4};
        
        \draw[white, very thick] (\halfw,0) -- (\halfw,\h);

        \draw[black, thick, #5] (0,0) rectangle (2*\halfw,\h);
    \end{tikzpicture}%
    }

    \scalebox{0.7}{
    \begin{tabular}{c|cccccc}
        \toprule
        \textbf{($a_i$$\downarrow$,$a_j$$\rightarrow$)} 
        & $a_1^J$ & $a_2^A$ & $a_3^A$ 
        & $a_4^A$ & $a_5^J$ & $a_6^J$ \\
        \midrule
        
        $a_2^A$ &
        \splitcell{cAtt}{.59}{cSup}{.20}{dashed} & 
        - & - & - & - & - \\
        
        $a_3^A$ &
        \splitcell{cAtt}{.96}{cAtt}{.20}{dashed} & 
        \splitcell{cSup}{.45}{cAtt}{.60}{solid} &  
        - & - & - & - \\
        
        $a_4^A$ &
        \splitcell{cAtt}{.91}{cAtt}{.20}{dashed} & 
        \splitcell{cSup}{.94}{cAtt}{.20}{solid} &  
        \splitcell{cSup}{.06}{cAtt}{.60}{solid} &  
        - & - & - \\
        
        ${a_5^J}$ &
        \splitcell{cSup}{0.98}{cSup}{1.}{solid} &  
        \splitcell{cAtt}{.49}{cNei}{N}{dashed} &  
        \splitcell{cAtt}{.91}{cAtt}{.20}{dashed} &   
        \splitcell{cNei}{N}{cNei}{N}{dashed} &   
        - & - \\
        
        ${a_6^J}$ &
        \splitcell{cSup}{.98}{cSup}{1.0}{solid} &  
        \splitcell{cNei}{N}{cNei}{N}{dashed} &   
        \splitcell{cAtt}{.77}{cNei}{N}{dashed} & 
        \splitcell{cNei}{N}{cNei}{N}{dashed} &   
        \splitcell{cSup}{.95}{cSup}{.34}{solid} & 
        - \\
        
        $a_7^J$ &
        \splitcell{cSup}{.97}{cSup}{1.}{solid} & 
        \splitcell{cNei}{N}{cAtt}{.20}{dashed} & 
        \splitcell{cAtt}{.83}{cAtt}{1.}{dashed} & 
        \splitcell{cAtt}{0.35}{cAtt}{.60}{dashed} & 
        \splitcell{cSup}{.93}{cSup}{.60}{solid} &  
        \splitcell{cSup}{.96}{cSup}{.60}{solid} \\
        
        \bottomrule
    \end{tabular}
    }
    \caption{Comparison between our approach (left) vs Prompt-based (right) for relation strength estimation on argument pairs $(a_i,a_j)$ in $D_1$, with $i>j$.  Red-/Green-colored cells refer to attacks/supports, while grey-shaded cells marked with `N' indicate the absence of a relation. Numeric values denote the relation strength. 
    Solid (resp., dashed) borders refer to pairs of arguments stated by the same author (resp., different authors). 
    }
    \label{tab:combined_heatmap_compact}\vspace{-6mm}
\end{table}

\noindent
\textbf{Relation Strength.}
We compare the proposed relationship strength estimation (cf. Equation~\ref{eq:prob_label_relation_diseq}) and its prompt-based counterpart. 
Their difference stands in the approach for computing $p_{i,j}^{r}$: the latter just prompts an LLM, whereas our approach also takes into account log-probabilities. 
In our experiments, the threshold $\vartheta$ has been set to $0$, intuitively stating that a relation type $x$ from $a_i$ to $a_j$ exists iff the model's confidence in $x$ exceeds the combined confidence in the other two types. 
Table~\ref{tab:combined_heatmap_compact} compares the two approaches over arguments in $D_1$. 
Qualitatively, the prompt-based approach tends to rely on surface-level lexical cues, often misclassifying argumentative relations. Clear examples are the attacks $(a_3^A,a_2^A)$, $(a_4^A,a_2^A)$, and $(a_4^A,a_3^A)$ between arguments {by} the same author (i.e. Arthur); and the support $(a_2^A,a_1^J)$.
Quantitatively, considering all debates in $\cal D$ and all possible relations, our approach led to the extraction of $45.60\%$ attack (resp., {$34.72\%$} support) relations, $7.17\%$ (resp., $1.08\%$) of which represent attacks {on arguments by} the same author (resp., supports {on arguments by} different authors. 
Conversely, the prompt-based approach led to the extraction of $32.64\%$ attack (resp., $45.60\%$ support) relations, 
$17.75\%$ (resp., $14.93\%$) of which represent attacks {on arguments by} the same author (resp., supports {on arguments by}   different authors.

\section{Reasoning in \nFAKB}

The semantics of an \nFAKB is based on the concept of \textit{fuzzy interpretations}.
A fuzzy interpretation $\mathcal{I} = (\Delta^\mathcal{I}, \cdot^\mathcal{I})$ provides a membership degree for objects belonging to the different concept and role names, that are interpreted as fuzzy unary and binary relations over $\Delta^\mathcal{I}$, respectively. 
Differently from classical KB (cf.  Section~\ref{sec:prel}),  $(i)$ names are taken from $\mathsf{N_A^+}$, $\mathsf{N_C^+}$, and $\mathsf{N_R^+}$, and 
$(ii)$ the codomain of $\cdot^\mathcal{I}$ is a number in $[0,1]$ (instead of $\{0,1\}$).
The function $\cdot^\mathcal{I}$ is extended to general concepts and roles as follows: 
$(i)$ $(\exists \textt{Q}(a))^\mathcal{I} = sup_{b \in \Delta} (\textt{Q}(a,b))^{\cal I}$; 
$(ii)$ $(\neg \boon{A}(a))^\mathcal{I} = 1 \!-\! (\boon{A}(a))^\mathcal{I}$;
$(iii)$ $(\textt{Q}^-(a,b))^\mathcal{I} \!=\! (\boon{Q}(b,a))^{\cal I}$;
$(iv)$ $(\neg \textt{Q}(a,b))^\mathcal{I} \!=\! 1 \!-\! (\textt{Q}(a,b))^\mathcal{I}$.

\vspace*{1mm}
\noindent 
\textbf{Consistency Checking.}
As in \DLL, inconsistency in \nFAKB (i.e., \nFAKB admitting no models) may only arise from negative inclusions (NIs), also called disjointness assertions. An example of NI is $A \sqsubseteq \neg B$, stating that, for every individual $a$ and interpretation $\cal I$, $(A(a))^{\cal I} + (B(a))^{\cal I} \leq 1$ must hold.
A consistency checking algorithm for KBs with fuzzy ABox has been proposed in \citep{MailisT14}. 
It adapts the {\textt{Consistent}} algorithm in \DLL~\citep{CalvaneseGLLR07}, and consists of the following main steps: 
\begin{enumerate}\itemsep=-2pt
\item
(Collect negative inclusions.)
Compute the closure of NIs to find all implied contradictions. For instance, $A \sqsubseteq B$ and $B \sqsubseteq \neg C$ imply $A \sqsubseteq \neg C$.
All such pairs $(X,Y)$ where $X \sqsubseteq \neg Y$ are collected.
\item
(Fire Boolean Queries.)
For every NI, e.g. $B_1 \sqsubseteq \neg B_2$, assuming that any fuzzy fact $\<f(a),v\>$ is stored as $f(a,v)$, if the result of the query $\exists x, v_1, v_2. B_1(x,v_1) \wedge B_2(x,v_2) \wedge v_1 > 1 - v_2$ is empty, the knowledge base is consistent.
\end{enumerate}

Clearly, one can employ any linear programming solver to handle this system of linear inequalities, guaranteeing that consistency checking in \nFAKB can be performed efficiently.

\vspace*{2mm}
\noindent 
\textbf{Querying (consistent) FABKs.}
Given a consistent \nFAKB in hand, an interesting problem is that of query answering. 
While in classical \nQBAF this would correspond to checking the acceptance status associated with a specific (goal) argument, in \nFAKB more general queries can be posed.

Given a \nFAKB $\FAKB=\tFAKB$, a \emph{fuzzy conjunctive query} (FCQ) on $\FAKB$ is an expression of the form: 
$\q(\vec{x}) = \exists \vec{y}.\, \varphi(\vec{x}, \vec{y}),$
where \( \varphi(\vec{x}, \vec{y}) \) is a conjunction of  atoms of the forms $\textt{A}(a)$ or $\textt{r}(a,b)$, where $\textt{A}\in \mathsf{N_C^+}$, $\textt{r}\in \mathsf{N_R^+}$, $a,b$ are variables or individuals from $\mathsf{N_I^+}$, $\vec{x}$ denotes the set of free variables, and $\vec{y}$ denotes the set of existentially quantified variables.

From the syntactic viewpoint, the difference between FCQs and  CQs stands in the language used, that is FCQs make use of the augmented sets $\mathsf{N_I^+}$, $\mathsf{N_C^+}$, and $\mathsf{N_R^+}$, thus allowing to express argumentative concepts and roles (e.g., $\textt{Arg}$, $\textt{att}$, and $\textt{sup}$).
A union of FCQs (UFCQ) is an expression of the form  
$\q(\vec{x}) = \q_1(\vec{x}) \vee \dots \vee \q_n(\vec{x}),$
where each $\q_i(\vec{x})$ is a FCQ.
An (U)FCQ $\q(\vec{x})$ is Boolean if all its variables are existentially quantified (thus $\vec{x}$ is empty).

As FCQ associates degrees to tuples in the answers, to compute degrees, we refer here to Zadeh's fuzzy logics \citep{Zadeh65}. That is, operators $\wedge$, $\vee$, and $\neg$, are respectively interpreted as: 
$(i)$ $(\varphi_1 \wedge \varphi_2)^{\cal I} = min(\varphi_1^{\cal I}, \varphi_2^{\cal I})$,  
$(ii)$ $(\varphi_1 \vee \varphi_2)^{\cal I} = max(\varphi_1^{\cal I}, \varphi_2^{\cal I})$, and 
$(iii)$ $(\neg \varphi)^{\cal I} = 1 -\varphi^{\cal I}$.

It is well known that the key property of \DLL\xspace is that any CQ can be rewritten into an equivalent UCQ using only the ABox.
Consider a \DLL\xspace KB $\tFAKB$, and let ${\cal I}_\ABox=(\Delta^{{\cal I}_\ABox},\cdot^{{\cal I}_\ABox})$ be the interpretation where $\Delta^{{\cal I}_\ABox}$ contains all constants in $\ABox$, and the codomain of $\cdot^{{\cal I}_\ABox}$ denotes the corresponding (boolean) value \wrt  $\ABox$. 
Moreover, let $\mathit{ans}(q(\vec{x}),\ABox)= \{ \vec{t} \mid (q(\vec{t}\ ))^{\cal I_A} > 0 \}$ denote the set of answers to the query $q(\vec{x})$ over the `database' $\ABox$,\footnote{For standard \DLL, it is equivalent to $\{ \vec{t} \mid (q(\vec{t}\ ))^{\cal I_A} = 1\}$. The motivation for adopting $>0$ stands, as stated next, that we make use of $\mathit{ans}(q(\vec{x}),\ABox)$ also when $\ABox$ is an FABox.} and $\mathit{cert}(q(\vec{x}),\tFAKB)$ denote the set of certain answers to the query $q(\vec{x})$ over the \DLL\xspace KB $\tFAKB$, i.e. the set of tuples contained in all models of $\tFAKB$. 
For \DLL\xspace there exists an algorithm $\mathsf{ref}$ that, given a CQ $q(\vec{x})$, gives as output an UCQ $\mathsf{ref}(q(\vec{x}),\TBox)$ such that $\mathit{cert}(q(\vec{x}),\tFAKB) = \mathit{ans}(\mathsf{ref}(q(\vec{x}),\TBox),\ABox)$.
That is, instead of computing the models of the KB $\tFAKB$ to answer a query $q(\vec{x})$, the query is rewritten into an equivalent one $\mathsf{ref}(q(\vec{x}),\TBox)$ which can be directly executed over $\ABox$.

The answer to an FCQ $q(\vec{x})$ on a \nFAKB $\FAKB=\tFAKB$ is a set of fuzzy tuples $\<\mathit{tuple},\mathit{degree}\>$, called ``\textit{fuzzy certain answers}'' and defined as follows:
{\begin{equation*}
\mathit{fcert}(q(\vec{x}),\FAKB) = 
\{ \<\vec{t},\alpha\> \mid \alpha = \inf_{{\cal I} \models \FAKB} (q(\vec{t}\ ))^{\cal I} \ {\wedge \ \alpha > 0} \}
\end{equation*}
}
\noindent
that is, $\alpha$ is the infimum of the degrees assigned to tuple $\vec{t}$ in all fuzzy {models of} \nFAKB. 
Analogously, the fuzzy answer to an FCQ $q(\vec{x})$ over a FABox $\ABox$ is:
{\setlength{\abovedisplayskip}{2pt}
\setlength{\belowdisplayskip}{2pt}
\begin{equation*}
{\mathit{fans}}(q(\vec{x}),\!\ABox) \!=\! 
\{ \<\vec{t},\alpha\> \!\mid\! \vec{t} \in \mathit{ans}(q(\vec{x}),\!\ABox) \wedge\, \alpha \!=\! (q(\vec{t}\ ))^{{\cal I}_\ABox} \}.
\end{equation*}
}

As stated next, also (U)FCQ are FO rewritable in \nFAKB.

\begin{proposition}\label{prop:fcert}
    Let $\FAKB$=$\tFAKB$ be a \nFAKB and $q(\vec{x})$ be an (U)FCQ. Then, it holds that:
    \begin{center}
    {\centering $\mathit{fcert}(q(\vec{x}),\FAKB) = \mathit{fans}(\mathsf{ref}(q(\vec{x}),\TBox),\ABox)$.}
\end{center}
\end{proposition}

We note that this result is in line with existing literature on fuzzy \DLL, where the FO rewriting algorithm $\mathsf{ref}$ for standard \DLL\xspace KBs has been adapted to cope with fuzzy KBs~\citep{MailisT14,PasiP23}.

As the answer to an FCQ may contain fuzzy tuples $\<\vec{t},(q(\vec{t}\ ))^{{\cal I}_\ABox}\>$ having a very low answer degree 
$(q(\vec{t}\ ))^{{\cal I}_\ABox}$, we assume that any FCQ is either of the form $q(\vec{x}) \geq k$, with $k \in (0,1]$ or of the form $q(\vec{x}) > k$, with $k \in [0,1)$, where $k$ is a user defined value, whose aim is to filter out those tuples where $(q(\vec{t}\ ))^{{\cal I}_\ABox} < k$ and $(q(\vec{t}\ ))^{{\cal I}_\ABox} \leq k$, respectively. 
We assume that $q(x)$ stands for $q(x) > 0$. 

\begin{example}\label{ex:query}\rm
Let $\FAKB$ be the  \nFAKB of Example~\ref{ex:FAKB-syntax}. We next report two possible FCQs $q_i(\vec{x})$ (with $i\in [1,2]$), their relative meaning and (fuzzy, certain) answers $\mathit{fcert}(q_i(\vec{x}),\FAKB)$.
\begin{itemize}
    \item Arguments with acceptance strength greater than $0.75$:\\ 
    $\mathit{fcert}\big((\textt{Arg}(x) >\! 0.75),\FAKB\big)=\{\<a_3,0.8\>\}$;
    \item The text of climate arguments from left-wing parties\\ attacked by arguments from right-wing ones:\\ $q_2(x)= \exists y,z,w.\ \ \textt{topic}(y,\textt{climate})\wedge \textt{textOf}(y,x)\wedge\\ \hspace*{\fill} \textt{Left}(z)\wedge \textt{att}(w,y) \wedge \textt{Right}(w)$\\ $\mathit{fcert}(q_2(x)>0,\FAKB)=\{\<\ta{1},0.703\>\}.$\hspace*{\fill}$\Box$
\end{itemize}
\end{example}

\noindent
As stated next, the (data) complexity suggests that fuzzy certain answers in \nFAKB can be efficiently computed.

\begin{proposition}\label{prop:answering}
    Let $\FAKB=\tFAKB$ be a consistent \nFAKB and $\q(\vec{x})$ an UFCQ on $\FAKB$. Then, $\mathit{fcert}(q(\vec{x}),\FAKB)$ can be computed in PTIME \wrt $|\TBox|$ and in LOGSPACE \wrt $|\ABox|$.
\end{proposition}

\section{Related Work}
Early work in Argument Mining (AM) relied on traditional NLP pipelines and supervised machine learning techniques~\citep{Palau11argmining,stab-gurevych-2014-identifying,stab2017persuasiveessay}.  
Advances in Transformer-based architectures such as BERT~\citep{mayer2020ecai}, and T5~\citep{kawarada-etal-2024-argument}, have significantly improved AM performance, due to the greater contextual awareness and ability to catch the argumentative flow in texts.  
Recently, the advent of generative LLMs demonstrated remarkable capabilities in AM tasks framed as text-generation~\citep{chen-etal-2024-exploring-potential}.
Empirical studies have shown that LLMs outperform previous approaches in identifying argumentative components, classifying their roles~\citep{favero2025leveraging}, and detecting attack and support relations~\citep{gorur-etal-2025-large}, even in low-resource settings~\citep{kashefi-etal-2023-argument}. 
Moreover, LLMs exhibit a remarkable ability to perform zero-shot or few-shot argumentation mining~\citep{cabessa2024context}, significantly reducing the need for costly manual annotations.
Other similar works explored neuro-symbolic approaches that use LLMs for constructing (QB)AFs~\citep{FreedmanAAAI25,gorur-etal-2025-large,cabessa-etal-2025-argument,mayer2020ecai}.

Fuzzy logic has been successfully applied in several engineering fields \citep{Mamdani74,Ross16}. Beyond engineering, it has been adopted in decision support and expert systems to facilitate approximate, human-like reasoning in complex environments \citep{DuboisPrade80,Zimmermann01}. 
In fuzzy logic programming~\citep{dubois1991fuzzy} such as the fuzzy variant of answer set programming (FASP), propositions are allowed to be graded~\citep{lukasiewicz2008fuzzy,van2007introduction,janssen2011foundations}. 
Although fuzziness in \nFAKB could alternatively be modeled via FASP, we adopt DLs since they offer a standardized formal foundation consistent with Semantic Web technologies and OWL standards, while ensuring interoperability with existing fuzzy ontologies.


\section{Conclusion and Future Work}
We presented \nFAKB, a framework integrating learning-based extraction with quantitative reasoning and ontology-mediated query answering. Unlike systems reliant on black-box prompting, our implementation derives initial argument and relation strengths directly from the transformer's output log-probabilities to capture internal model confidence. By embedding this data into an ontological structure, \nFAKB enables both consistency checking and graded conjunctive query answering. 
This approach broadens the scope of argumentative reasoning from simple acceptability checks to complex queries on the entire extracted knowledge, allowing for more appropriate explainability of decisions and conclusions.
Although we have referred to UFCQ as a query language, narrow SQL can be used to express practical queries operating on texts.
Future work will focus on investigating incremental techniques for the construction of  {\nFAKB}s, so that debates can be efficiently analyzed in real time.

\section*{Acknowledgements}
\noindent
We acknowledge financial support from PNRR MUR projects FAIR (PE0000013) and SERICS (PE00000014),  project Tech4You (ECS0000009),  and (MUR) PRIN 2022 project S-PIC4CHU (2022XERWK9).

\clearpage

\bibliographystyle{abbrv}
\bibliography{refs}

\clearpage

\appendix

\onecolumn
\section*{Appendix}

\centering
\scalebox{0.95}{
\begin{tcolorbox}[
    title={Debate $D_1^*$ on climate change.},
    colback=white,
    colframe=black,
    fonttitle=\bfseries,
    arc=0pt, outer arc=0pt,
    boxrule=0.5pt,
    left=0pt, right=0pt, top=0pt, bottom=0pt,
    sharp corners
]
\centering
\begin{minipage}{0.49\linewidth}
    \includegraphics[width=\linewidth]{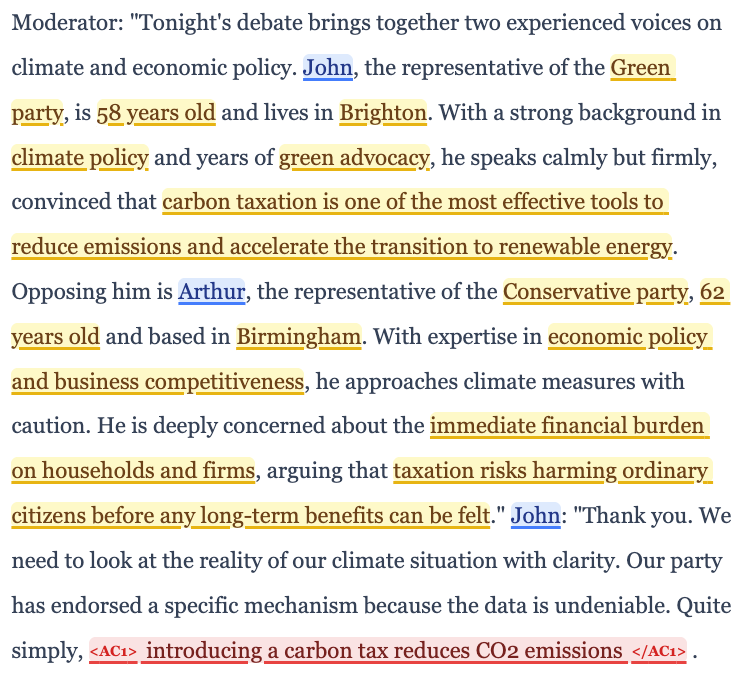}
\end{minipage}
\hfill
\begin{minipage}{0.49\linewidth}
    \includegraphics[width=\linewidth]{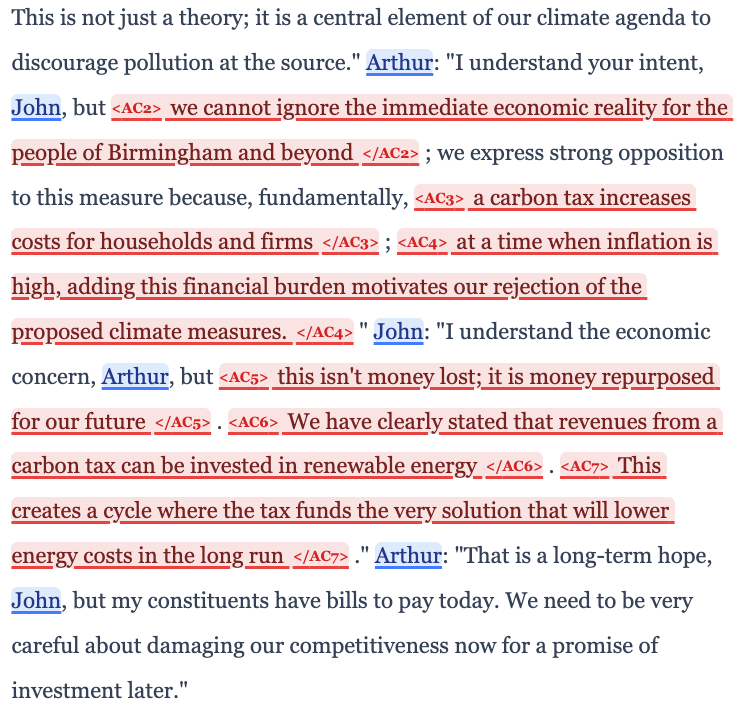}
\end{minipage}
\end{tcolorbox}
}

\begin{figure*}[h]
    \centering
    \includegraphics[width=0.8\linewidth]{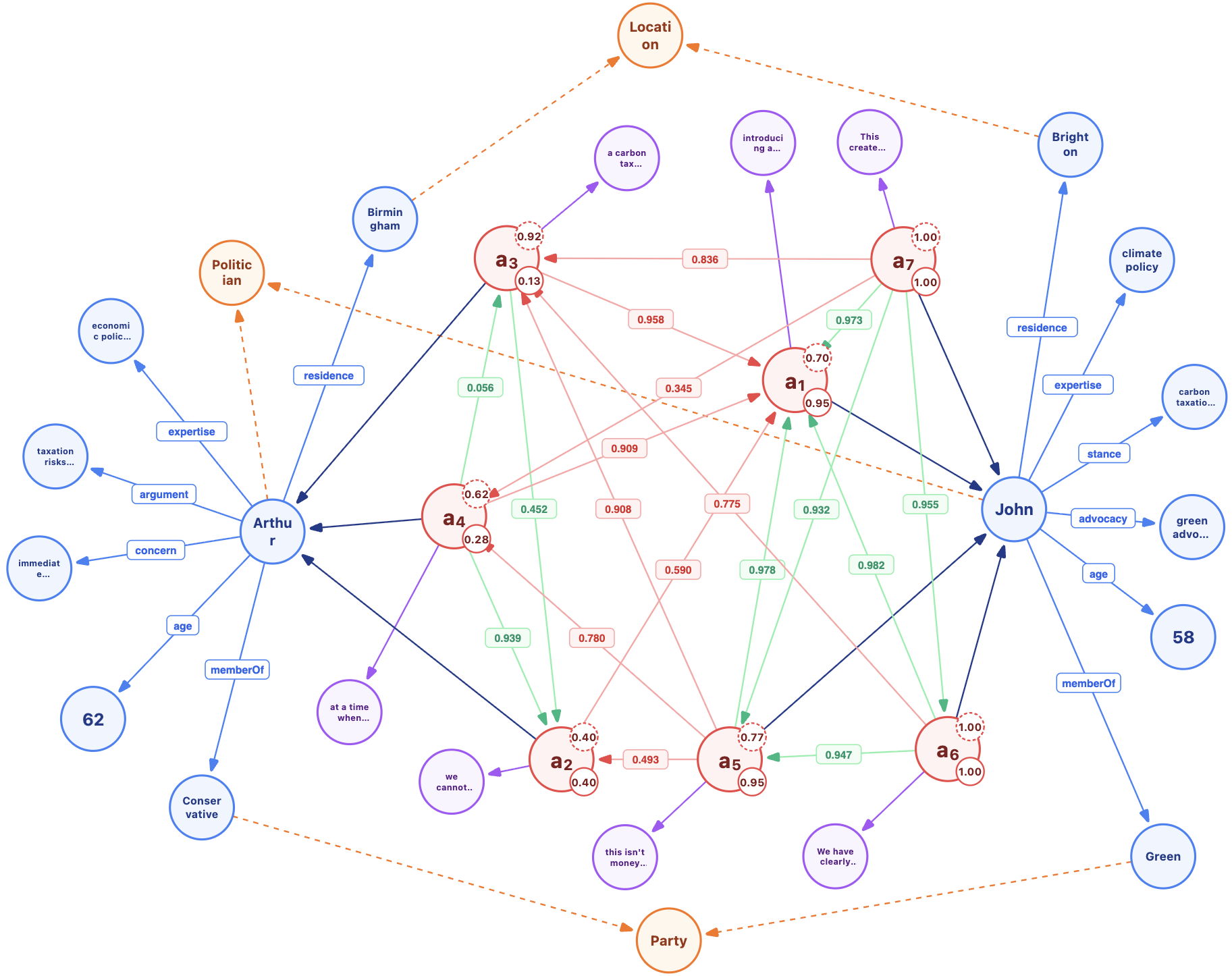}
    \caption{An example of updated FAbox, visualized in the form of a knowledge graph, and extracted from an extended version of debate $D$ of Example~\ref{ex:intro0}. Nodes are color-coded by type: violet for text of arguments, blue for entities, red for argument identifiers, and yellow for concepts. Similarly, edges represent specific relations: green/red denotes support/attacks, while dark blue, light blue, and yellow represent authorship, role instances, and concept instances, respectively. Upper/lower numbers within each argument denote its initial/updated strength.}
    \label{fig:step3-app}
\end{figure*}

\end{document}